\documentclass{egpubl}
\usepackage{eg2025}
 
\ConferencePaper        % uncomment for (final) Conference Paper
\CGFccby
\usepackage[T1]{fontenc}
\usepackage{dfadobe}  

\biberVersion
\BibtexOrBiblatex
\usepackage[backend=bibtex,bibstyle=EG,citestyle=alphabetic,backref=true]{biblatex} 
\electronicVersion
\PrintedOrElectronic

\ifpdf \usepackage[pdftex]{graphicx} \pdfcompresslevel=9
\else \usepackage[dvips]{graphicx} \fi

\usepackage{egweblnk} 
\usepackage{lineno}
\usepackage{amsmath}
\usepackage{amssymb}

\newcommand{\yang}[1]{{\color{black}#1}}

\newcommand{\ie}{\textit{i}.\textit{e}.} 

\title[StyleBlend: Enhancing Style-Specific Content Creation in Text-to-Image Diffusion Models]%
      {StyleBlend: Enhancing Style-Specific Content Creation in Text-to-Image Diffusion Models}

\author[Chen et al.]
{\parbox{\textwidth}{\centering Zichong Chen
        \quad Shijin Wang \quad Yang Zhou\thanks{Corresponding author.} %\orcid{0000-0001-5923-423X}
        }
        \\
{\parbox{\textwidth}{\centering Visual Computing Research Center, CSSE, Shenzhen University
       }
}
}
\begin{document}

\teaser{
 \includegraphics[width=0.9\linewidth]{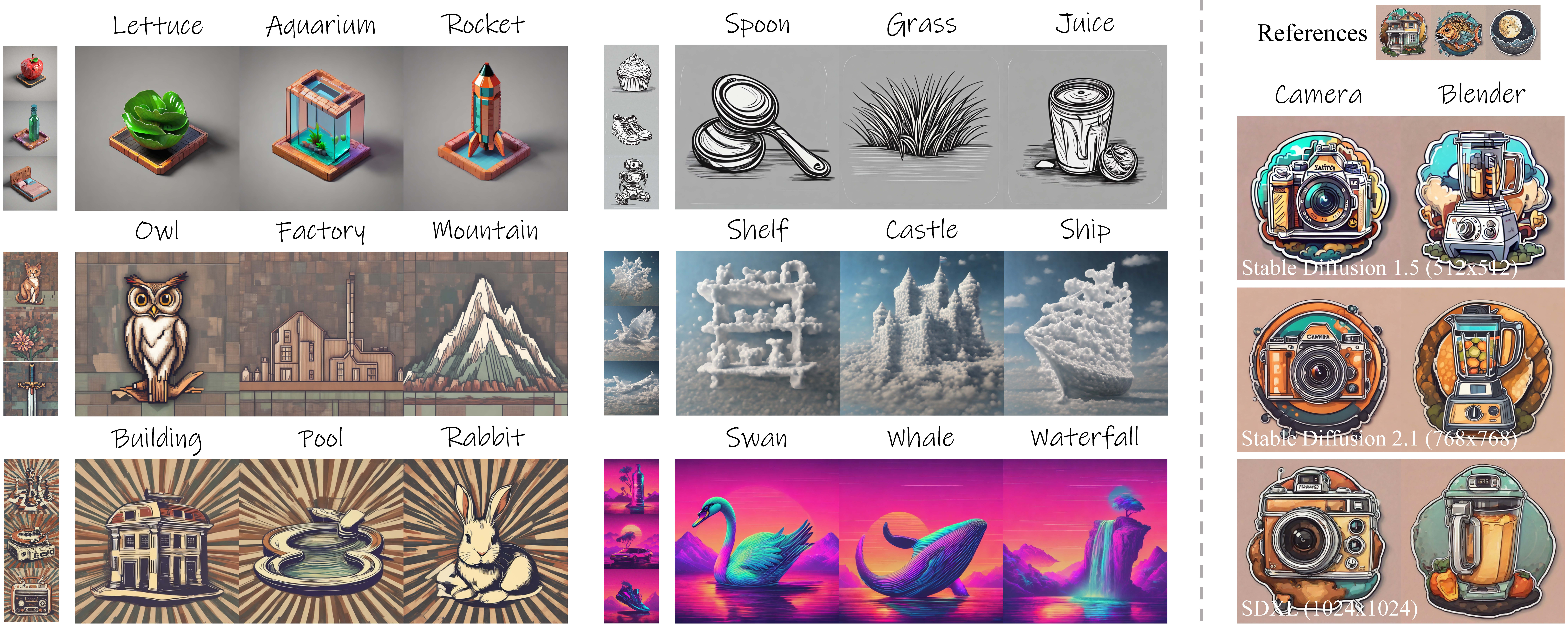}
 \centering
  \caption{\textbf{Style-specific text-to-image generation.} Given a limited set of style images arranged vertically on the left side, our method generates content images that align with the text semantics while also exhibiting satisfactory stylistic effects (left). Additionally, our approach can be applied to the Stable Diffusion series without any degeneracy or incompatibility (right).}
\label{fig:teaser}
}

\maketitle
%-------------------------------------------------------------------------
\begin{abstract}
   Synthesizing visually impressive images that seamlessly align both text prompts and specific artistic styles remains a significant challenge in Text-to-Image (T2I) diffusion models. This paper introduces StyleBlend, a method designed to learn and apply style representations from a limited set of reference images, enabling content synthesis of both text-aligned and stylistically coherent. Our approach uniquely decomposes style into two components, composition and texture, each learned through different strategies. We then leverage two synthesis branches, each focusing on a corresponding style component, to facilitate effective style blending through shared features without affecting content generation. StyleBlend addresses the common issues of text misalignment and weak style representation that previous methods have struggled with. Extensive qualitative and quantitative comparisons demonstrate the superiority of our approach.
%-------------------------------------------------------------------------



\end{abstract}  
%-------------------------------------------------------------------------
\section{Introduction}
In recent years, Text-to-Image (T2I) diffusion models have made remarkable progress in generating high-quality images based on textual prompts. Models like DALL·E \cite{ramesh2021dalle} and Stable Diffusion \cite{esser2024sd3, Podell2023SDXLIL, Rombach2021HighResolutionIS} have shown immense potential in various fields, including artistic creation. However, one of the remaining challenges is synthesizing images that seamlessly integrate textual prompts with specific artistic styles derived from a set of images. This integration is crucial for applications that demand a high level of style alignment, such as personalized art creation.

\begin{figure}[t]
    \centering
    \includegraphics[width=0.95\linewidth]{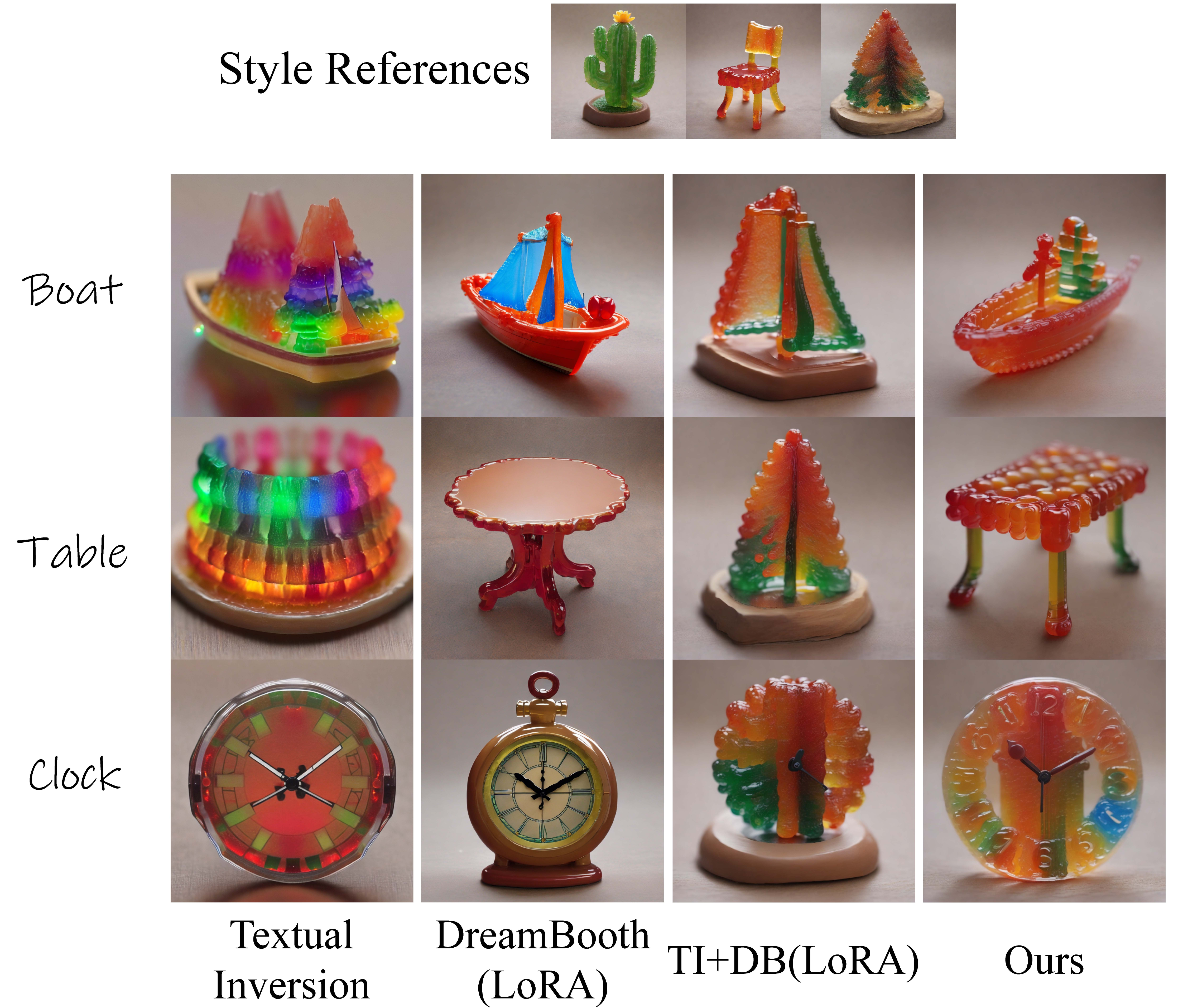}
    % \vspace*{-3mm}
        \caption{
    % \scriptsize
    \textbf{Motivation of StyleBlend.} Given the three reference images at the top, we leverage existing methods to learn style representations and generate results conditioned by the class-specific prompts. The widely-used techniques, such as Textual Inversion (TI) \cite{Gal2022AnII} and DreamBooth (DB) \cite{Ruiz2022DreamBoothFT}, struggle to produce high-quality style-specific images that maintain style coherence and text alignment simultaneously. We also experimented with combining these two approaches. While yielding surprising stylistic effects, it fell into overfitting. Our approach builds upon these observations and achieves much higher quality. Note that we only train additional LoRAs instead of the entire denoising network in DreamBooth for efficiency.
    }
    \label{fig:motivation}
    % \vspace*{-2mm}
\end{figure}

Current methods \cite{Gal2022AnII, kumari2022customdiffusion, Ruiz2022DreamBoothFT, Ruiz2024hyperdreambooth, alaluf2023neti, zhang2023prospect} typically finetune a pretrained T2I model over a set of images that share the same style, but often fall short in two crucial aspects: text alignment and style representation. Most of them struggle to balance style and content in the generated images, especially when there is very limited reference data. As shown in Fig. \ref{fig:motivation}, Textual Inversion (TI) \cite{Gal2022AnII} fails to reproduce the fine details of the given style and meanwhile suffers from poor semantic content alignment. DreamBooth (DB) retains semantic integrity much better but doesn't capture the style info precisely. Surprisingly, combining TI and DB yields notable stylistic effects in mimicking the references' appearance and texture, but overfitting issues occur; see the 3rd column of Fig.~\ref{fig:motivation} for example. We attribute this to the strong correlation between appearance and semantics, particularly under the challenge of data scarcity (limited data and lack of diversity). Overfitting enhances style representation but meanwhile sacrifices semantic alignment. This gap highlights the need for a method that can robustly blend text-aligned content with stylistic coherence, especially in few-shot cases.

To address these challenges, we propose StyleBlend, a method that decomposes the image style into two components: composition and texture. 
% Composition focuses on semantic structure and layout info without sacrificing semantic integrity. 
\yang{Composition refers to the semantic structure and layout information of a scene, focusing on the arrangement and relationships between objects and elements without sacrificing semantic integrity. In contrast, texture emphasizes finer details and local appearance.}
% Texture emphasizes finer details and local appearance. 
To generate style-specific images, StyleBlend employs a dual-branch framework, where each branch corresponds to one style component.
To learn the style components separately, we start from our insights on the two simple yet widely-used techniques \cite{Gal2022AnII, Ruiz2022DreamBoothFT}. As discussed above, narrowing the gap between the generated domain and the target domain can significantly enhance the capture of style information. We thus learn the texture style of images using this approach. Then, we introduce a newly designed strategy upon SDEdit~\cite{meng2022sdedit} technique to capture the composition style, which can provide accurate semantic content with correct layout information in inference. Based on the two learned style representations, we blend them through {mutual} feature injection between the branches during synthesis. Qualitative and quantitative results show that, compared with existing methods, StyleBlend overcomes the common issues of text misalignment and weak style representation, offering a more robust solution for style-specific content generation in T2I diffusion models.

Our contributions are summarized as follows:
\begin{itemize}
    \item We introduce a new method for decomposing and learning style representations, enabling the synthesis of images that are both text-aligned and stylistically coherent.
    \item We develop a dual-branch synthesis framework with style features blending, achieving high-quality stylized image generation.
    \item We demonstrate the superiority of our approach through extensive qualitative and quantitative comparisons with existing state-of-the-art methods.
\end{itemize}

\section{Related work}

\subsection{T2I diffusion models and personalization}
Recent advancements in T2I diffusion models, particularly within the Stable Diffusion series \cite{esser2024sd3, Podell2023SDXLIL,Rombach2021HighResolutionIS}, have greatly improved the quality and diversity of images generated from text prompts. Building on this progress, a new task has emerged, called personalization, which tries to learn and generate new images of a specific subject based on a few reference images while preserving the key characteristics of that subject. Some approaches \cite{alaluf2023neti, Gal2022AnII, voynov2024p} introduce new tokens paired with corresponding textual embeddings to capture the subject. These extended embeddings are optimized by traditional diffusion loss and used to prompt the model in inference to generate desired results. Some other methods \cite{hao2024vico, hua2023dreamtuner, kumari2022customdiffusion, marjit2023diffusekrona, pang2024attndreambooth, Ruiz2022DreamBoothFT, Ruiz2024hyperdreambooth, shah2023ZipLoRA} focus on refining layer weights or integrating additional features to enhance the model's generative capabilities for personalized characteristics. \yang{Recently, inspired by inversion encoders in GAN~\cite{GAN_Inversion_Bau2019, psp_richardson2021, Pivotal_Roich2022}, image adapters have also been introduced for concept learning in T2I personalization~\cite{e4t2023, ye2023ip-adapter}.} However, the subject or concept of an image is usually more abstract and unspecific than style. As a result, these methods often struggle to maintain consistent text alignment and fully capture fine details of styles, particularly when data is limited.

\begin{figure*}[htb]
  \centering
  \includegraphics[width=0.95\linewidth]{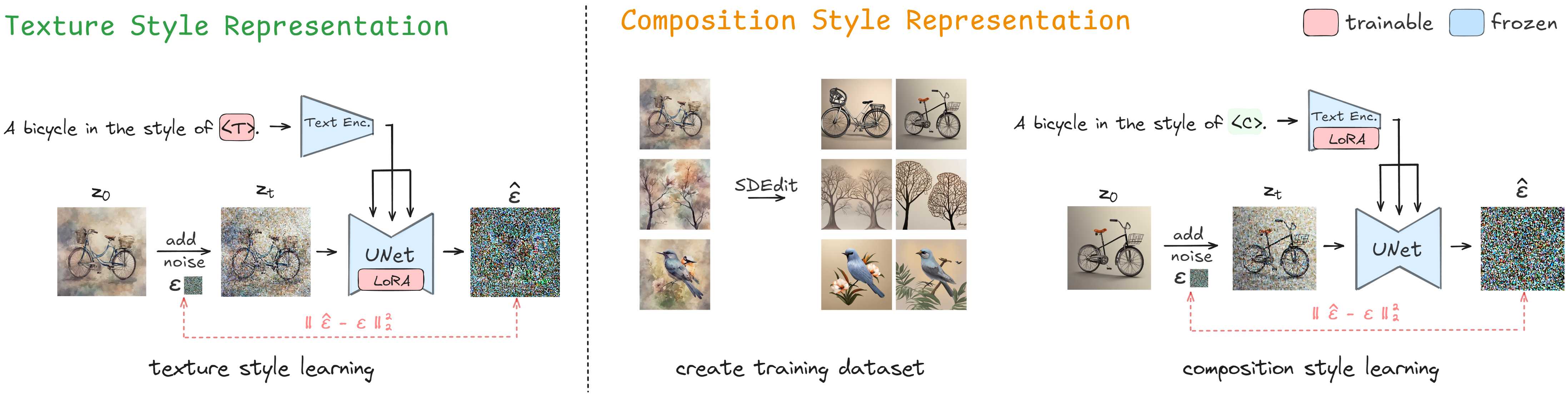}
  \caption{\label{fig:learning}
           \textbf{Style representation and learning in StyleBlend.} We decompose the image style into two components: texture style and composition style. Texture style refers to the texture and appearance of images. We first optimize {a global} embedding with an identifier ``<T>'' to {learn the subject of the given images, which can} narrow the generated domain, followed by training the LoRA weights for {finer-detailed} style learning (left). For composition style, we generate a set of training data using SDEdit \cite{meng2022sdedit} and then train the LoRA weights of the text encoder with an identifier ``<C>'' to capture the semantic structure and layout of images (right).
           }
\end{figure*}

\subsection{Style-specific image generation} 
Starting with the traditional style transfer \cite{gaty2016styletransfer}, which applies the artistic style of one image to the content of another, recent works \cite{chung2024styleid, he2024freestyle, wang2023stylediffusion, zhang2023inst, deng2024zstar} have advanced by incorporating diffusion models. Style-specific text-to-image generation, a notable subfield of personalization, aims to produce images that not only align with the provided text prompts but also embody specific artistic styles. Recent research in this area can be categorized into three main types: fine-tuning, training-free, and pretrained methods. Fine-tuning methods \cite{ahn2024dreamstyler, everaert2023diffusioninstyle, frenkel2024implicit, Gal2022AnII, jones2024customizingtexttoimagemodelssingle, kumari2022customdiffusion, Ruiz2022DreamBoothFT, sohn2023styledrop, frenkel2024implicit}, similar to the personalization approaches discussed above, learn the target style by fine-tuning T2I models on a given set of reference images that share the same style. Training-free methods, such as StyleAligned \cite{hertz2024stylealigned}, RB-Modulation \cite{rout2024rbmodulation}, Visual Style Prompting~\cite{jeong2024visual} and InstantStyle \cite{wang2024instantstyle}, leverage the prior knowledge learned by pretrained T2I models to integrate styles without the need for additional training. Pretrained methods \cite{chen2023controlstyle, gao2024styleshot, wang2024styleadapter, xing2024csgo, ye2023ip-adapter} are designed to handle arbitrary styles {based on large-scale datasets}, offering a versatile solution for style-specific image generation. However, existing fine-tuning methods still encounter the same issues as personalization approaches, while the training-free and pretrained methods struggle with capturing the precise style information defined by a given set of style images. Instead, we developed a new fine-tuning approach {that can vividly reproduce the style (both global and local)} while mitigating text misalignment issues found in existing personalization methods.

\section{Preliminaries}

\subsection{Diffusion models and personalization}
Diffusion models \cite{sohl2015deep, ho2020denoising} simulate a diffusion process where images are progressively perturbed with noise and then reconstructed through a learned denoising mechanism. Our approach builds upon the prominent Stable Diffusion models \cite{Podell2023SDXLIL,Rombach2021HighResolutionIS}, which performs the diffusion process in latent space and effectively utilize a U-Net architecture comprising multiple convolution and transformer \cite{vaswani2017attention} blocks as the diffuser.

Specifically, during the training of Stable Diffusion, an image $x$ is first compressed into latent space using a pretrained encoder $\mathcal{E}$. Then, noise is added to latent code $z$, and a denoising U-Net $\epsilon_\theta(\cdot)$ is trained to predict the added noise with the following objective:
\begin{equation}
\mathcal{L}_{\mathrm{diff}}=\mathbb{E}_{z\sim \mathcal{E}(x),y,\epsilon\sim\mathcal{N}(0,\textbf{\yang{I}}),t}{\big[}||\epsilon-\epsilon_\theta(z_t,t,c(y))||^2_2{\big]},
\label{eq:diff_loss}
\end{equation}
where $t$ denotes the timestep, and $y$ represents the textual condition, which is typically encoded into text embeddings using a CLIP encoder $c(\cdot)$ \cite{radford2021clip}.

There are two widely used techniques to personalize pretrained T2I models based on a set of images that share the same concept, Textual Inversion~\cite{Gal2022AnII} and DreamBooth~\cite{Ruiz2022DreamBoothFT}. Textual Inversion optimizes a new token with a corresponding embedding vector $e^*$ using the same objective as Eq. \ref{eq:diff_loss}. DreamBooth modifies the parameters throughout the entire denoising network, also with the same objective. Additionally, we can introduce LoRAs~\cite{hu2021lora} into specific layers of pre-trained diffusion networks. A finetuned LoRA works the same as DreamBooth but is more memory efficient.
See the first two columns of Fig. \ref{fig:motivation} for examples of style-specific T2I generation using these two approaches.

\subsection{Feature injection in T2I diffusion models}
The features in the convolutional layers and attention blocks of the denoising U-Net
have demonstrated their effectiveness in information encoding \cite{pnpDiffusion2022, cao2023masactrl, hertz2024stylealigned, alaluf2024cross}. In this work, we focus on the U-Net self-attention features, which are calculated according to the self-attention mechanism: % contained in the U-Net
\begin{equation}
\mathrm{Self\text{-}Attention}(Q,K,V)=\mathrm{softmax}{\big(}\frac{QK^T}{\sqrt{d}}{\big)}\cdot V,
\label{eq:attention}
\end{equation}
where \yang{$d$ is the feature dimension,} Q, K, and V $\in \mathbb{R}^{s\times d}$ represent the Query, Key, and Value, respectively. The Query here has been empirically shown to effectively capture semantic structures, while the Key and Value components excel at representing texture and appearance. Feature injection typically refers to the copying or concatenation of QKV features across synthesis branches within the same layers.

\section{StyleBlend}
Given a limited set of images with category-specific text prompts that share the same style, our goal is to synthesize {new} images using T2I diffusion models that align the image semantics with user-provided text prompts while maintaining style coherence with the given style references. In this section, we introduce the proposed StyleBlend method. We first present the representation and learning of image styles in Sec. \ref{sec:represent}, followed by an illustration of the inference process using our dual-branch synthesis framework with feature blending in Sec. \ref{sec:dualbranch}.

\subsection{Style representation and learning}
\label{sec:represent}
\begin{center}
    \textit{Image Style = Composition Style + Texture Style}
\end{center}

\noindent The style of an image encompasses various elements arranged hierarchically. Based on empirical observations, we define the formulation above, decomposing image style into two key components: composition and texture. Composition style captures the semantic structure and layout of an image, while texture style focuses on the finer details of texture and appearance. To learn these two style representations, we employ distinct training strategies tailored to each component, as depicted in Fig. \ref{fig:learning}. 

\textbf{Texture style representation.}
The key to our task is to learn a robust style representation, where existing methods, as shown in Fig. \ref{fig:motivation}, often fall short. Our key insight is that narrowing the gap between the generated style domain and the target domain at the outset can improve style transfer. For example, we can manually write a detailed style description, such as ``gelatin candy toy'', to serve as an identifier for the target style. However, this approach often struggles to precisely match the desired style, and obtaining such a description can be very challenging for users. 

We, therefore, adopt Textual Inversion~\cite{Gal2022AnII} to optimize a global embedding to narrow the domain gap. As illustrated on the left side of Fig. \ref{fig:learning}, we use a new token ``\textit{<T>}'' with a corresponding embeddings vector $e_T$ as a global style identifier, and optimize it with the following objective:
\begin{equation}
e_T=\mathrm{arg}\min_{e}\mathbb{E}_{z,y,e,t}{\big[}||\epsilon-\epsilon_\theta(z_t,t,c(y,e))||^2_2{\big]}.
\label{eq:ts_obj1}
\end{equation}
Afterward, we introduce LoRAs into the denoising network, which is trained similarly to Dreambooth \cite{Ruiz2022DreamBoothFT} to capture the finer details of style. We optimize the LoRA weights \yang{$\theta^T_{LoRA}$} using the same reference images with the same objective, defined as:
\begin{equation}
\yang{\theta^T_{LoRA}}=\mathrm{arg}\min_{\theta_{LoRA}}\mathbb{E}_{z,y,e_T,t}{\big[}||\epsilon-\epsilon_{\theta, \theta_{LoRA}}(z_t,t,c(y,e_T))||^2_2{\big]}.
\label{eq:ts_obj2}
\end{equation}

However, directly synthesizing results with this approach is prone to overfitting, as shown in the third column of Fig. \ref{fig:motivation}, yet it produces strong stylistic effects. Therefore, we design to extract the learned texture style here and further develop a separate process to capture the composition style, ensuring the preservation of semantic structure and layout.  

\textbf{Composition style representation.}
The composition style emphasizes semantic structure and layout rather than texture. To effectively capture the compositional information, we start by generating a set of images using an image editing tool \cite{meng2022sdedit}, as depicted in the middle of Fig. \ref{fig:learning}. We corrupt a reference image by adding noise to a certain extent and then denoise the noisy image back to a clear state with simple text prompts from its class name, e.g., ``\textit{A bicycle.}'' for an image containing a bicycle. Such generated images share highly similar structures and layouts with reference images while introducing variations in texture style. 

We learn the composition style solely from the generated dataset. Unlike the texture style representation, here we add LoRAs to the text encoder, as illustrated on the right side of Fig.~\ref{fig:learning}. Additionally, We introduce a new token ``\textit{<C>}'' with the corresponding embedding vector $e_C$ to identify the composition style and train the LoRA weights \yang{$\theta^C_{LoRA}$} with the following objective:
\begin{equation}
\yang{\theta^C_{LoRA}}=\mathrm{arg}\min_{\theta_{LoRA}}\mathbb{E}_{z,y,e_C,t}{\big[}||\epsilon-\epsilon_{\theta}(z_t,t,c_{ \theta_{LoRA}}(y,e_C))||^2_2{\big]}.
\label{eq:cs_obj}
\end{equation}

Note that we empirically found that directly training the LoRAs without optimizing $e_C$ can effectively prevent overfitting and, in advance, yield better performance. Therefore, we only optimize the LoRA weights of the text encoder to capture the composition style.

\begin{figure}[t]
  \centering
  \includegraphics[width=0.95\linewidth]{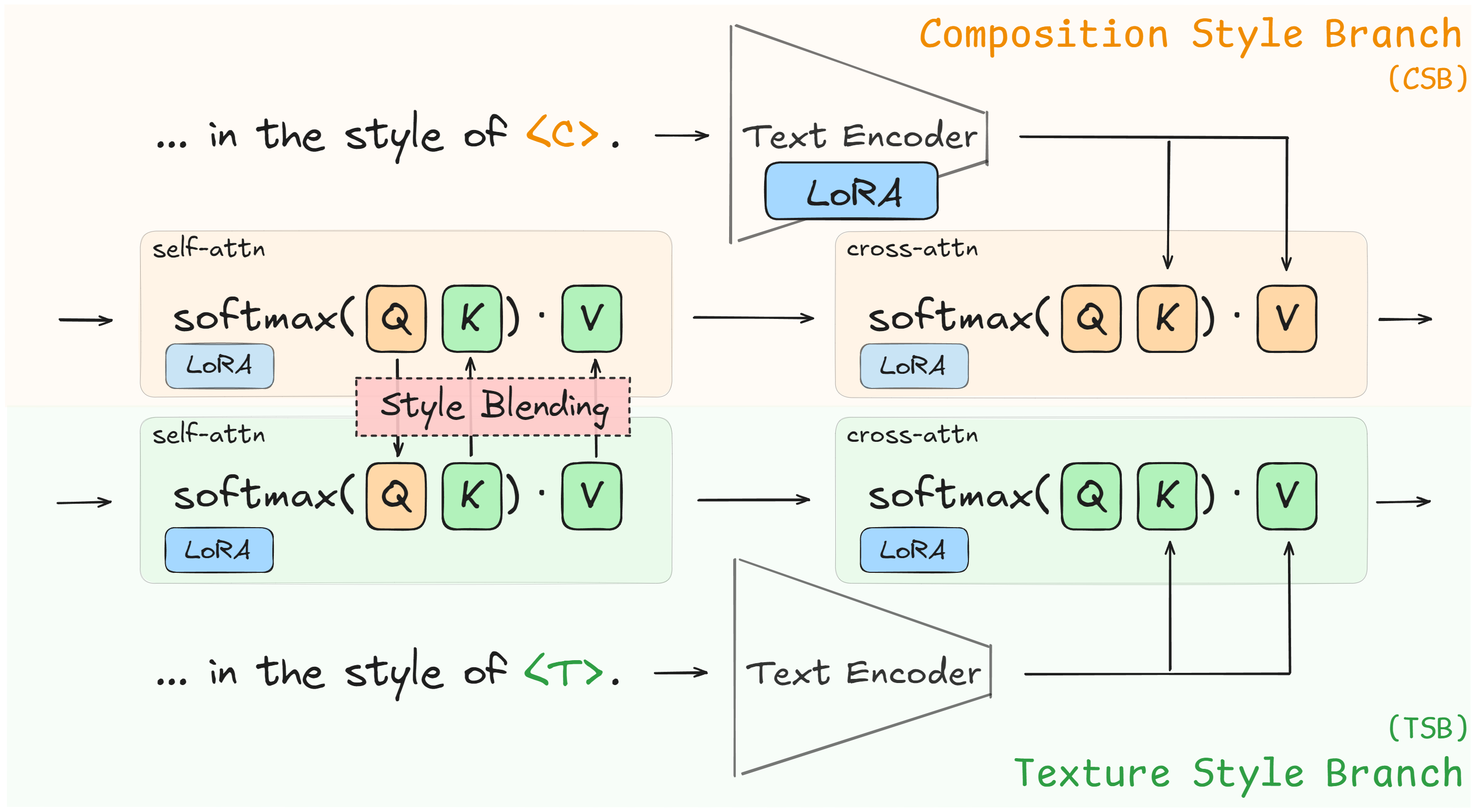}
  \caption{\label{fig:inference}
           \textbf{Dual-branch synthesis framework for inference.} The framework comprises two branches: a composition style branch (CSB) and a texture style branch (TSB), which produce composition style and texture style, respectively. We employ {mutual} feature injection to blend these style components. In the injected features, the Q features from CSB represent the composition style, while the KV features from TSB describe the texture style.
           }
\end{figure}

\begin{figure*}[t]
    \centering
    \includegraphics[width=1.\linewidth]{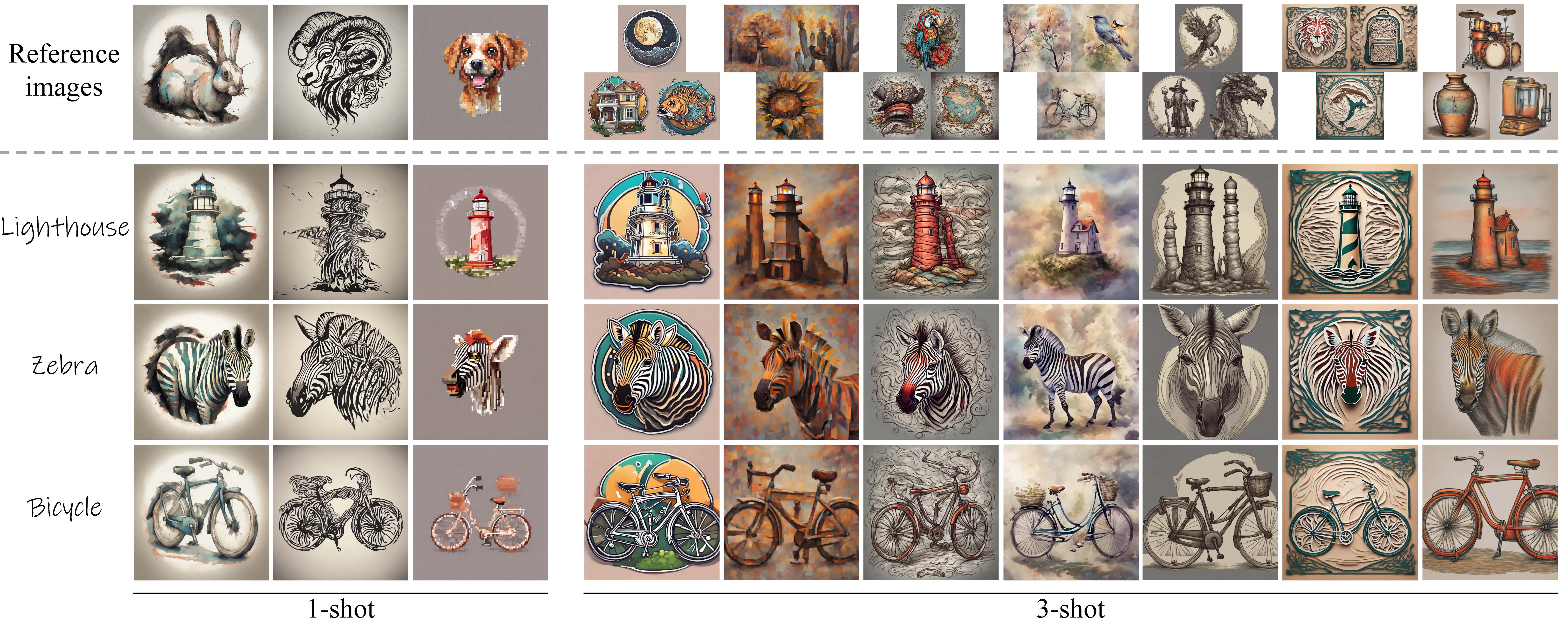}
    % \vspace*{-3mm}
        \caption{
    % \scriptsize
    Gallery of StyleBlend results for 1-shot (left) and 3-shot (right) cases, where style references are shown at the top.
    }
    \label{fig:gallery}
    % \vspace*{-2mm}
\end{figure*}

\subsection{Features blending across style components}
\label{sec:dualbranch}
After style learning, we obtain two sets of parameters \yang{($(\theta_{LoRA}^{C}, e_C)$ and $(\theta_{LoRA}^{T}, e_T)$)}: one for synthesizing results with composition style and the other for texture style. To generate results that match the desired style effects, we need to effectively ``add'' these two distinct styles following our formulation outlined above.

As illustrated in Fig. \ref{fig:inference}, we construct a dual-branch synthesis framework to separately generate composition and texture styles. Inspired by recent works \cite{pnpDiffusion2022, cao2023masactrl, hertz2024stylealigned} on feature manipulations in T2I models, we fetch the QKV features from the self-attention blocks and blend these features across the two synthesis branches for effective style integration. Empirical findings suggest that Q features capture semantic structure and layout while K and V represent texture and appearance. Accordingly, we swap QKV features between the two branches: the Q features from the texture style branch (TSB) are replaced with those from the composition style branch (CSB), while the KV features from CSB are replaced with those from TSB. This ensures proper blending of composition and texture in the final output, which is taken from the texture style branch. We refer to this approach as StyleBlend.

\section{Experiments}

\subsection{Implementation details}
\label{sec:implement}
We have implemented our method over the Stable Diffusion series (v1.5, v2.1, SDXL) \cite{Rombach2021HighResolutionIS,Podell2023SDXLIL} \yang{on a single RTX 4090. The learning processes for composition and texture styles are independent, with no specific priority assigned to their training order. We initialize $e_T$ with the token embedding of ``artistic'' and $e_C$ with a rare word such as the exclamation mark.} 
During training, we apply early stopping after a few hundred steps, which is enough to get satisfactory results. In practice, we typically optimize 500 steps for style embeddings and LoRA weights during texture style learning, while only 300 steps are required to optimize the LoRA weights of the text encoder during composition style learning. We set the learning rate to 0.0001 for optimizing all LoRA weights, using a larger learning rate of 0.01 for the style embeddings. \yang{The rank of LoRAs is set to 16 across all our experiments.} To generate a plausible dataset for composition style learning, we synthesize 10 images for each reference image. \yang{We apply noise perturbation to the reference images at 0.8T for Stable Diffusion v1.5 and v2.1, and 0.9T for SDXL, serving as the initial code for SDEdit sampling. The sampling process involves 20 linear steps.} %

\yang{For the computation time, training on Stable Diffusion 2.1 at a resolution of 768x768 takes approximately 75 seconds for dataset creation, 40 seconds for composition style learning, and 2 minutes for texture style learning. Inference requires around 4 seconds per image. Unless specified, the results presented are generated using Stable Diffusion v2.1 as the foundation model.}  %Our experiments are conducted on a single RTX 4090. The learning processes for composition and texture styles are independent, with no specific priority assigned to their training order.

\begin{figure*}[t]
    \centering
    \includegraphics[width=1.0\linewidth]{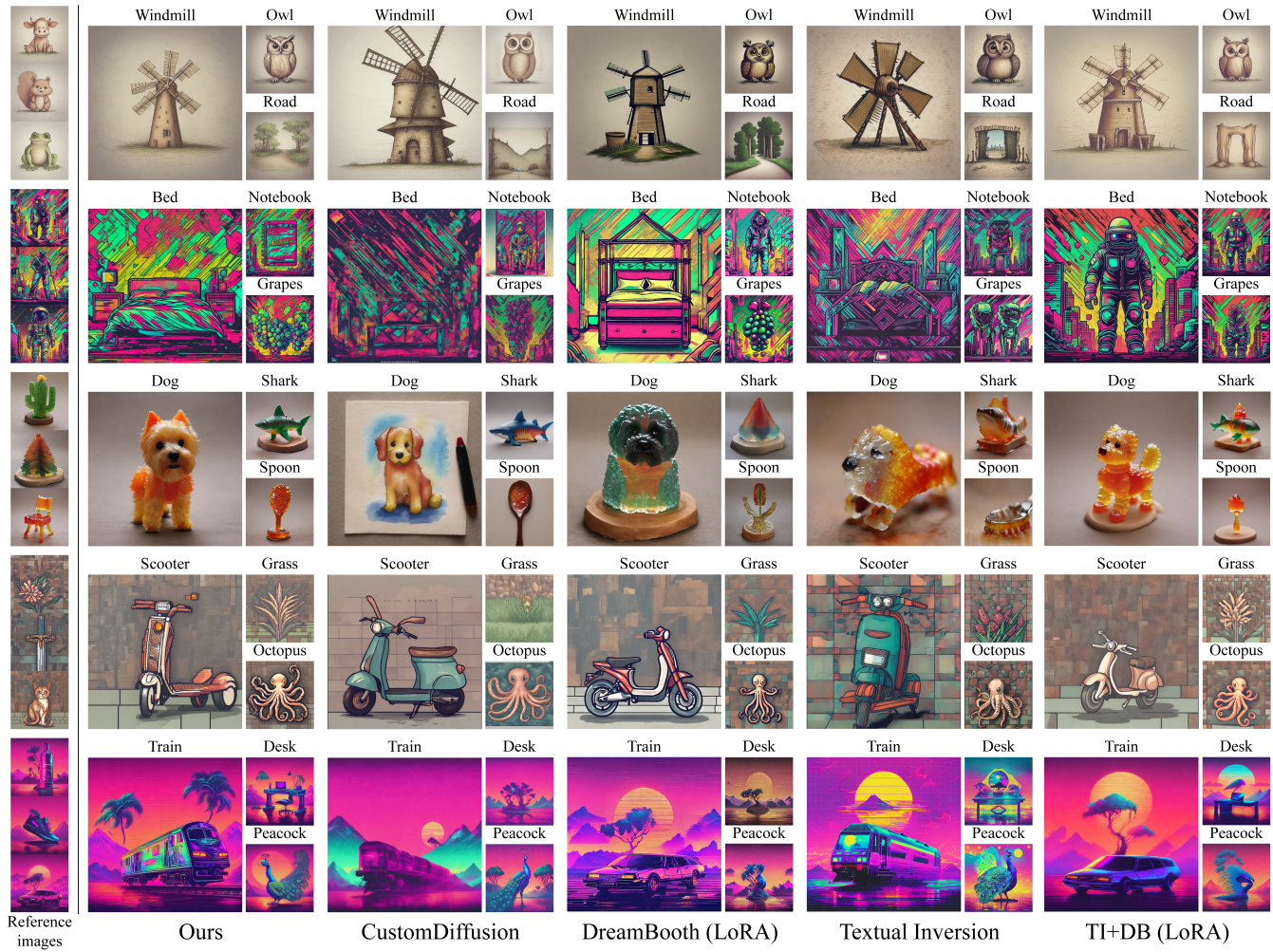}
    % \vspace*{-3mm}
        \caption{
    % \scriptsize
    Qualitative comparison to finetuning-based methods for style-specific T2I generation in 3-shot cases.}
    \label{fig:3shot}
    % \vspace*{-2mm}
\end{figure*}

\begin{figure}[t]
    \centering
    \includegraphics[width=1.\linewidth]{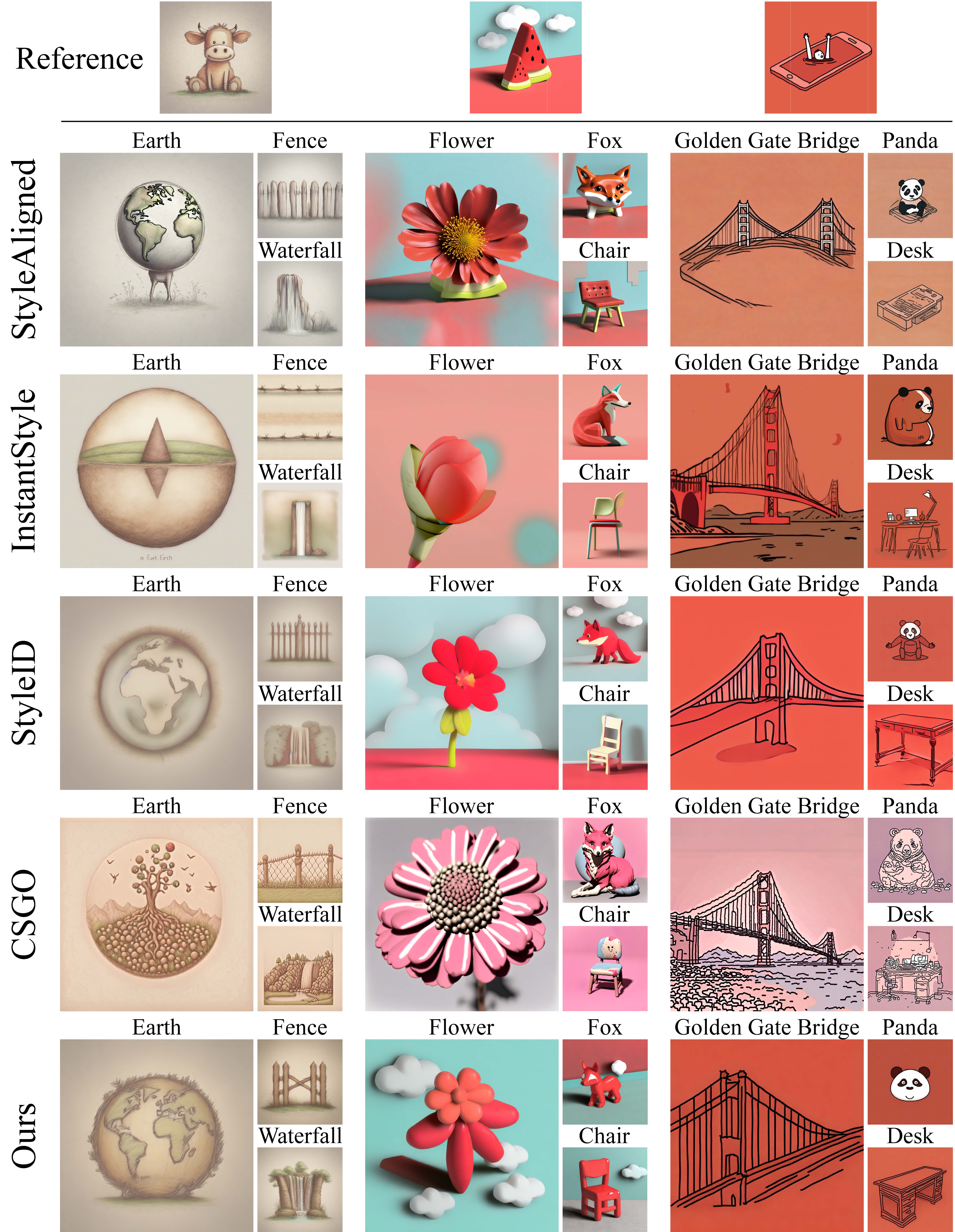}
    % \vspace*{-3mm}
        \caption{
    % \scriptsize
    \yang{Qualitative comparison to existing 1-shot methods in style-specific T2I generation, where CSGO~\cite{xing2024csgo} is a model pretrained on large style dataset, and the other three (StyleAligned~\cite{hertz2024stylealigned}, InstantStyle~\cite{wang2024instantstyle}, and StyleID~\cite{chung2024styleid}) are training-free methods.}}
    \label{fig:1shot}
    % \vspace*{-3mm}
\end{figure}

\begin{figure}[htb]
    \centering
    \includegraphics[width=1.\linewidth]{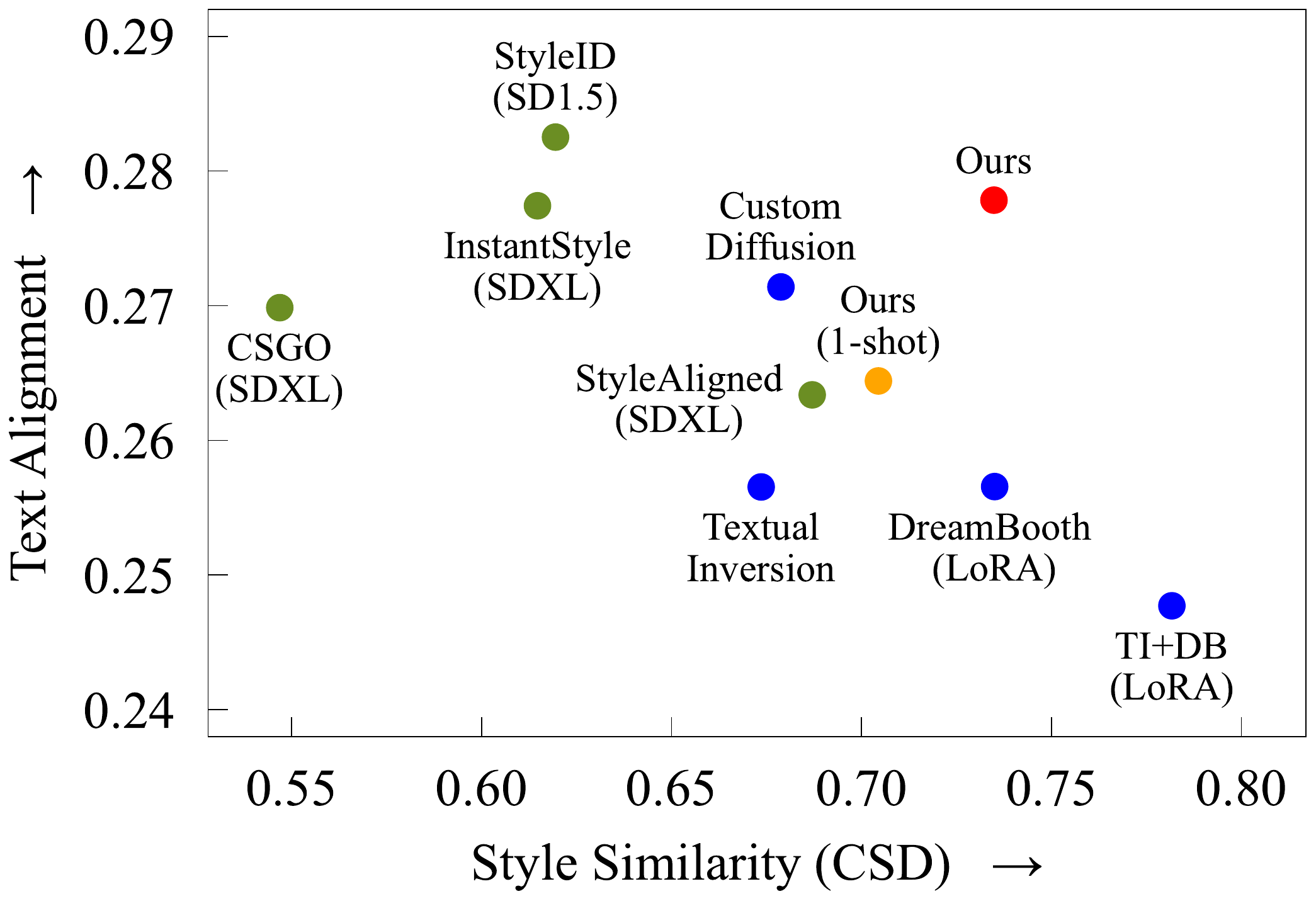}
    % \vspace*{-2mm}
        \caption{
    % \scriptsize
    \yang{Quantitative evaluation. The finetuning-based methods trained with three reference images (3-shot cases) are marked as blue scatter points, the pretrained and training-free approaches that infer from a single image (1-shot cases) are shown in green. Our results for the 3-shot and 1-shot cases are indicated by red and orange scatter points, respectively.}
    }
    \label{fig:comp_score}
    % \vspace*{-2mm}
\end{figure}

\textbf{Dataset.} To test and evaluate our approach, we collected several style images from the Internet, and synthesized additional sets of images that share the same style using StyleAligned \cite{hertz2024stylealigned} and VSP~\cite{jeong2024visual} over SDXL \cite{Podell2023SDXLIL}. We focus on 3-shot and 1-shot cases for fair comparison. We format the text prompts with the same templates:\textit{ ``A ... in the style of <C>''} for composition style and \textit{``A ... in the style of <T>''} for texture style, where the ellipses are filled with class-specific prompts, \ie, the class names.

\textbf{Metrics.} We utilize the Contrastive Style Descriptor (CSD) \cite{somepalli2024csd} to access style similarity to the target style, and the CLIP-Score \cite{radford2021clip} to evaluate text alignment with the input text prompts. Specifically, we measure the style similarity by computing the cosine similarity of style embeddings extracted by CSD between the generated images and the target style image. For the style set with multiple reference images, we first calculate the score between the generated image and each reference image and then take the maximum one as the final score for each generated image. The CLIP-Score is computed by measuring the cosine similarity between the CLIP embeddings of generated images and the corresponding input text prompts.

\subsection{Results and comparisons}
\textbf{Style-specific T2I synthesis.} Fig. \ref{fig:teaser} and Fig. \ref{fig:gallery} showcase the results of StyleBlend applied to various style examples. We focus on few-shot cases involving three reference images (3-shot) or a single image (1-shot). The results demonstrate that our method effectively manages both style coherence and text alignment simultaneously.

\textbf{Comparison.} 
To validate the effectiveness of our approach, we conducted qualitative and quantitative comparisons to several baselines. These include finetuning-based methods such as DreamBooth \cite{Ruiz2022DreamBoothFT}, Textual Inversion \cite{Gal2022AnII}, and CustomDiffusion \cite{kumari2022customdiffusion}, 
\yang{as well as a pretrained model CSGO~\cite{xing2024csgo} and three training-free methods: StyleAligned~\cite{hertz2024stylealigned}, InstantStyle~\cite{wang2024instantstyle}, and StyleID~\cite{chung2024styleid}.} For the baselines, we trained DreamBooth and Textual Inversion for 1000 steps using a learning rate of 0.0005 and 0.05, respectively. For other methods, we adhered to the configurations released with the open-source repositories. \yang{To run StyleAligned, we use GPT to generate a detailed style description for the DDIM inversion of the reference image and style generation. Furthermore, since StyleID requires a content image as input, we first generate content images using our CSB branch to provide a well-structured layout, and then apply StyleID to transfer the reference style to the generated images.} We also compared our approach with a variant that combines Textual Inversion and DreamBooth. Note that we optimize a set of LoRAs for DreamBooth instead of the entire denoising network.

Figs.~\ref{fig:3shot} and~\ref{fig:1shot} present the qualitative comparisons. Specifically, we compare our approach to finetuning-based methods in the 3-shot case in Fig.~\ref{fig:3shot}. We can see that recent advances suffer from weak style representation: the generated styles are not coherent with the target styles. Some of these methods even exhibit text misalignment, as seen in the 2nd and 5th rows, where the image content does not correspond accurately to the given text prompts. In contrast, our method effectively addresses these shortcomings, surpassing the baselines both in style and text alignment, leading to high-quality results. In Fig.~\ref{fig:1shot}, we compare our method with \yang{the pretrained model and training-free methods} in the 1-shot case, given that these methods accept only a single style image as input. Although the baselines perform well in terms of text alignment, they also struggle with style representation, much like the finetuning-based methods. \yang{Moreover, we found StyleAligned produced some destroyed structures, which we infer was caused by content leakage. For StyleID, the generated content images greatly affect the final results, resulting in less stylization than ours.}

Fig.~\ref{fig:comp_score} further presents quantitative comparisons on the metrics mentioned above, i.e., the CSD-based style similarity and the CLIP-Score for text alignment. 
% The finetuning-based methods trained with three reference images are marked as blue scatter points, while the pretrained approaches that infer from a single image are shown in green. Our results for the 3-shot and 1-shot cases are indicated by red and orange scatter points, respectively.
All the scores are computed across the five styles presented in Fig.~\ref{fig:3shot}. From the scores, our method achieves higher values for both style coherence and semantics alignment in the 3-shot case. Although our 1-shot results exhibit a slight weakness in text alignment due to the extremely limited data, they still produce satisfactory style effects compared to many other methods. In contrast, most baselines struggle to balance style coherence and text alignment.

\begin{figure}[t]
    \centering
    \includegraphics[width=1.0\linewidth]{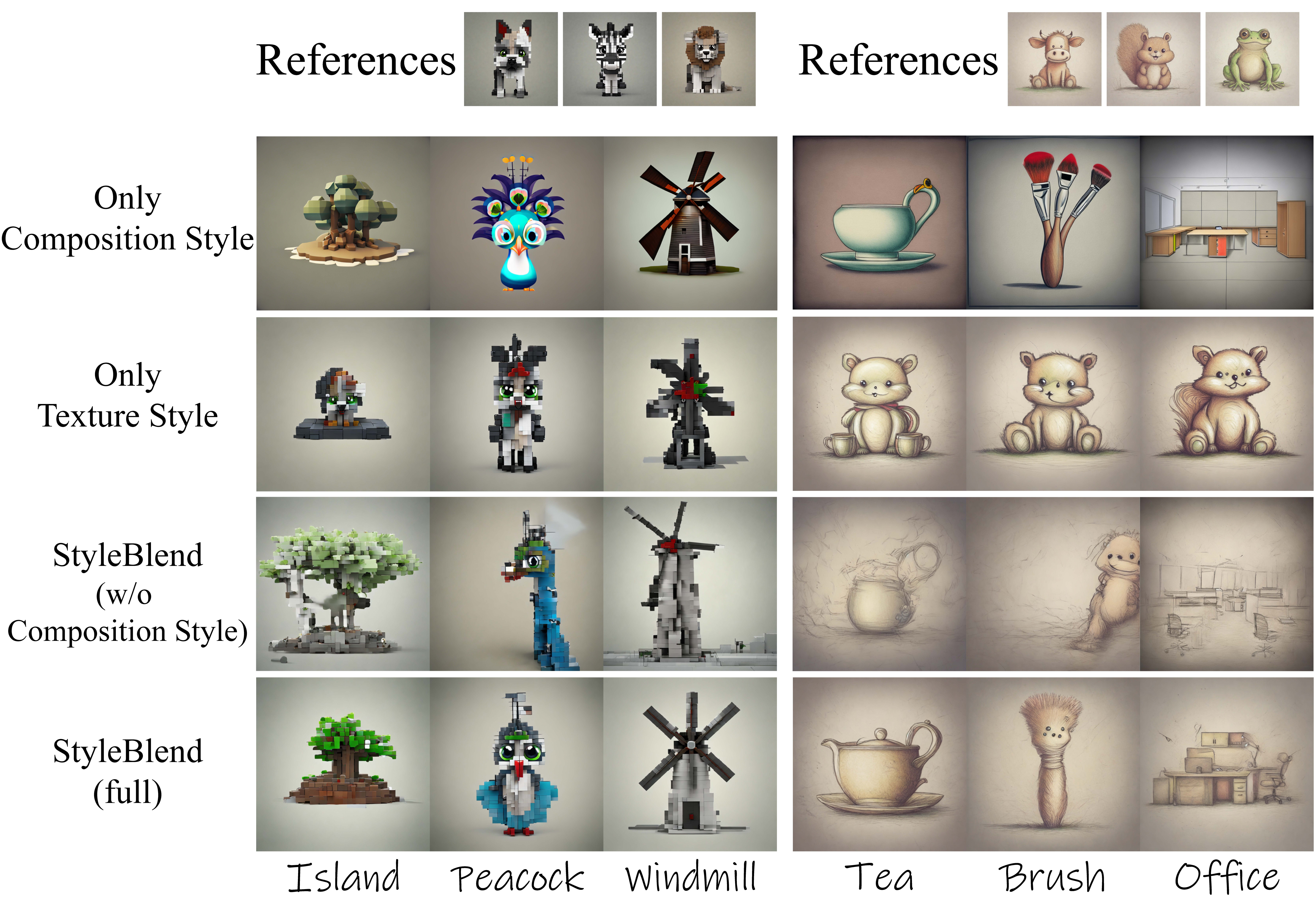}
    % \vspace*{-3mm}
        \caption{
    % \scriptsize
    Ablation of the style representations.}
    \label{fig:abl_representation}
    % \vspace*{-2mm}
\end{figure}

\begin{figure}[t]
    \centering
    \includegraphics[width=1.0\linewidth]{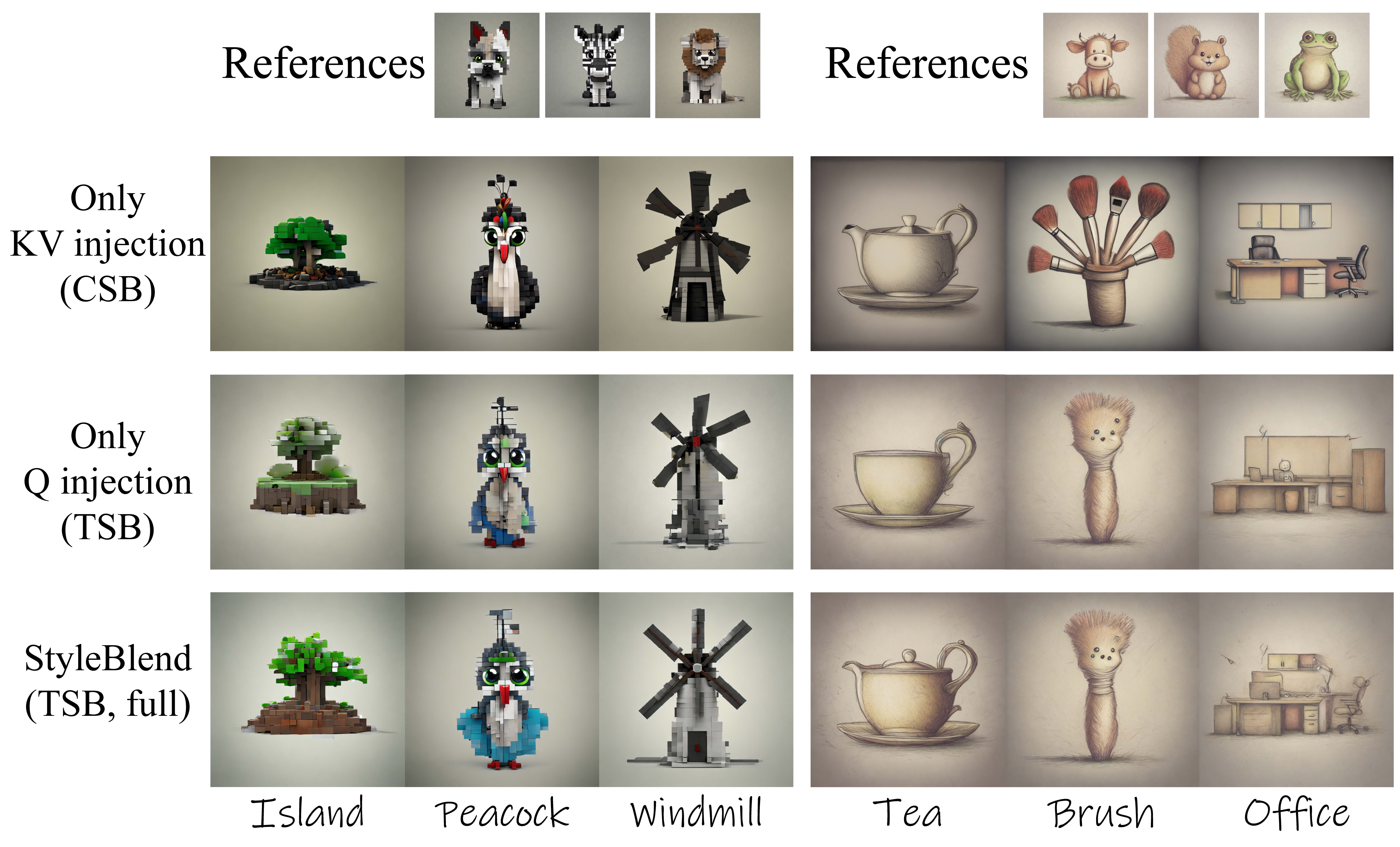}
    % \vspace*{-3mm}
        \caption{
    % \scriptsize
    Ablation of the feature injection.}
    \label{fig:abl_injection}
    % \vspace*{-2mm}
\end{figure}

\begin{figure}[t]
    \centering
    \includegraphics[width=1.\linewidth]{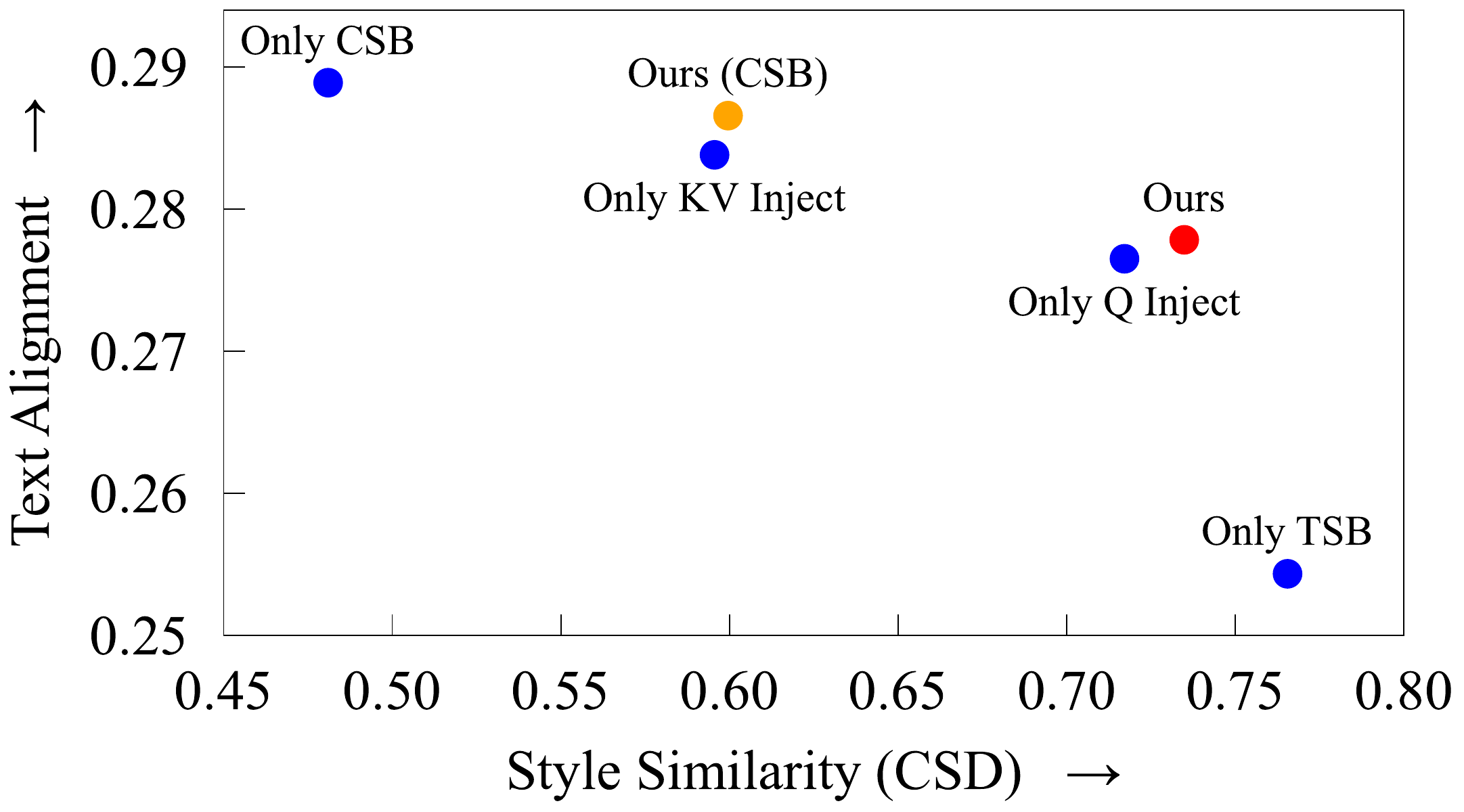}
    % \vspace*{-3mm}
        \caption{
    % \scriptsize
    \yang{Quantitative results of ablation study. Note Ours~(CSB) refers to the results produced by the composition style branch (CSB) of our full model, which is significantly inferior in style to the TSB results (Ours in the figure). We thus choose the outputs of TSB branch as our final results.}}
    \label{fig:abl_score}
    % \vspace*{-2mm}
\end{figure}

\begin{figure}[t]
    \centering
    \includegraphics[width=1.0\linewidth]{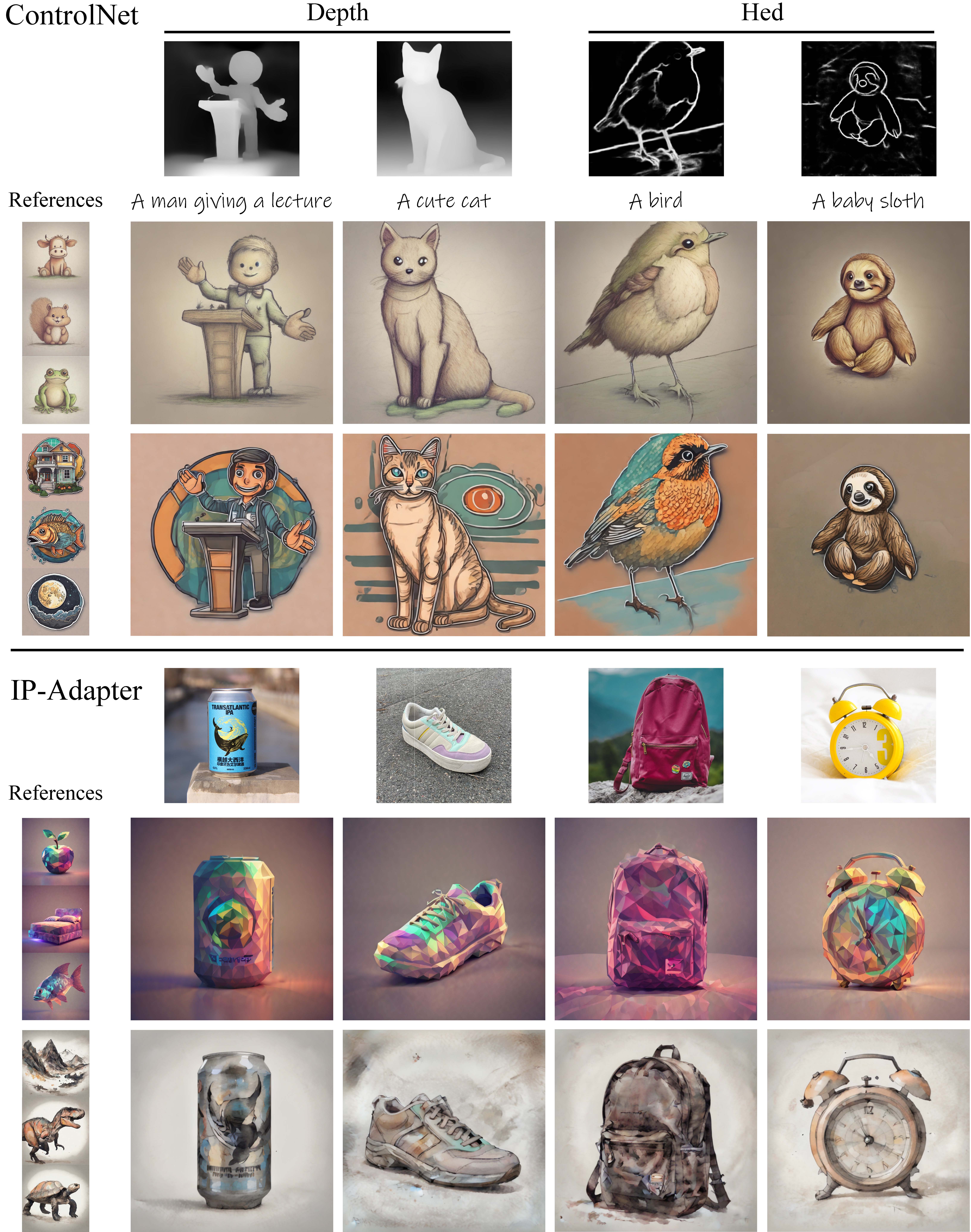}
    % \vspace*{-3mm}
        \caption{
    % \scriptsize
    Applying StyleBlend with ControlNet~\cite{zhang2023controlnet} and IP-Adapter~\cite{ye2023ip-adapter}.}
    \label{fig:application}
    % \vspace*{-2mm}
\end{figure}

\subsection{Ablation study}
We report the ablation results of StyleBlend on two aspects: style representation and feature injection. 

\textbf{Style representation.} As shown in Fig.~\ref{fig:abl_representation}, we separately synthesize results using only the composition style branch (CSB) and the texture style branch (TSB) (the first two rows). The results demonstrate that the composition style effectively captures the semantic structure and layout of images without compromising semantic integrity, while the texture style excels in representing texture and appearance but loses nearly all semantic information. \yang{Quantitative ablation results in Fig.~\ref{fig:abl_score} clearly illustrate the distinct performances of the composition and texture styles. Given that the TSB results significantly surpass CSB in style coherence, we choose the outputs from TSB as our final results.}
We also present the results of StyleBlend without \yang{learning the composition style from SDEdit data} to highlight the importance of semantic structure and layout in synthesis (the 3rd row). When the \yang{learned} composition style is removed--by excluding the learned LoRAs from the text encoder--the generated layout becomes disorganized, which further emphasizes the importance of composition style in capturing structure and layout information and preserving semantic coherence.

\textbf{Feature injection.} 
In the dual-branch synthesis framework, we extract the Q and KV features from the self-attention layers in the denoising U-Net to represent the composition and texture style features, respectively. \yang{Figs.~\ref{fig:abl_injection} and~\ref{fig:abl_score} show the qualitative and quantitative ablation study, respectively, on feature injection.}  When only KV features from the texture style branch (TSB) are injected into the composition style branch (CSB), the style of the results is apparently getting weak. On the other hand, injecting only Q features from the composition branch into the texture branch introduces some artifacts in the details, which indicates that running the two branches separately might cause incompatibility between the features of the two branches. In contrast, our full method, which leverages both branches and uses mutual feature injection, produces higher-quality results from the texture style branch.

\subsection{Additional results and applications}

\begin{figure}[t]
    \centering
    \includegraphics[width=1.0\linewidth]{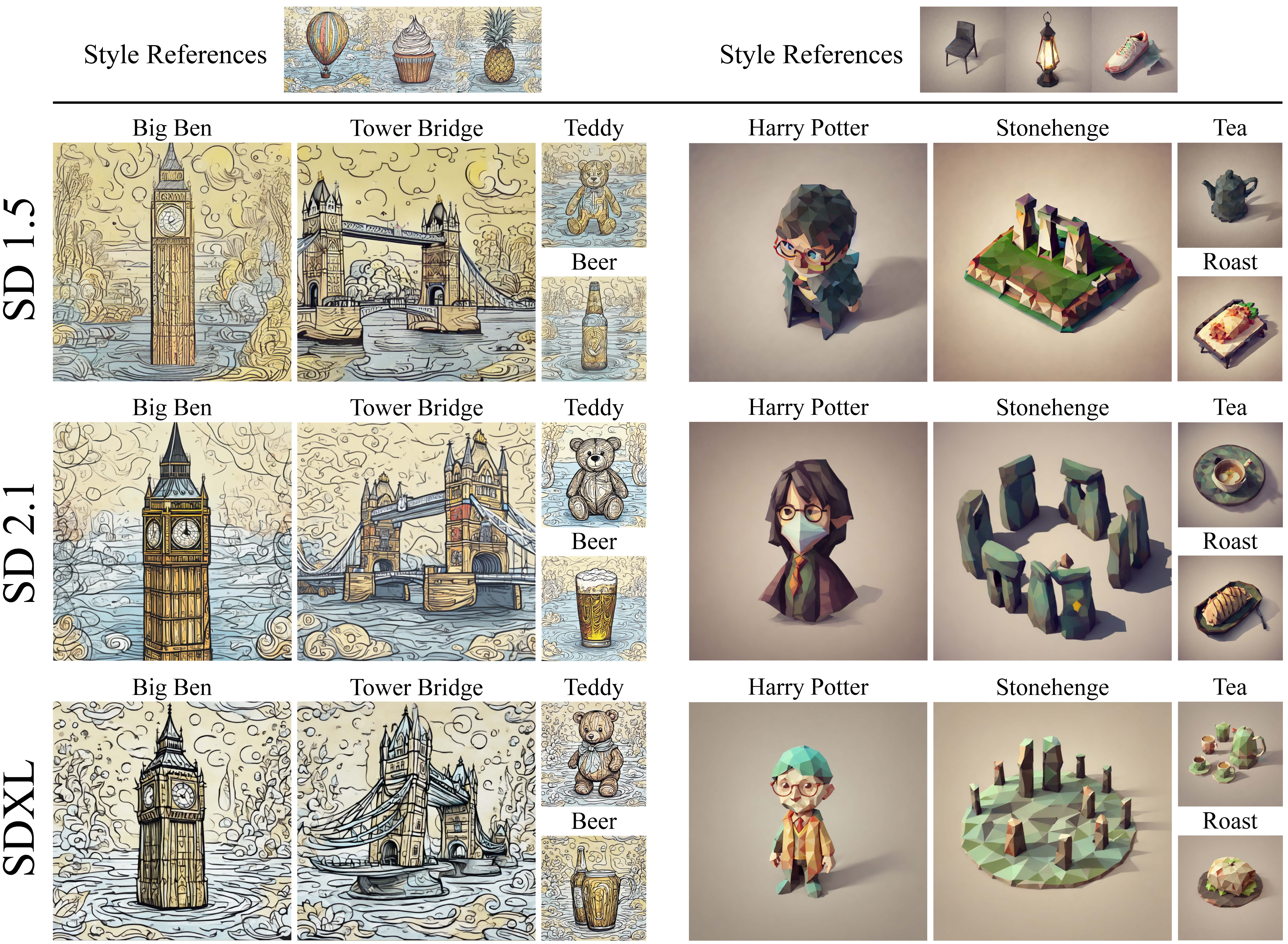}
    % \vspace*{-3mm}
        \caption{
    % \scriptsize
    Creative StyleBlend results obtained with various Stable Diffusion base models.}
    \label{fig:variations}
    % \vspace*{-2mm}
\end{figure}

\textbf{StyleBlend with ControlNet and IP-Adapter.}
Our method builds on pretrained SD models without modifying the original architecture or parameters. As a result, it can be easily integrated with many other pretrained plug-in methods, leveraging the extensive capabilities of the Stable Diffusion community. Fig.~\ref{fig:application} presents the results of StyleBlend combined with ControlNet~\cite{zhang2023controlnet} and IP-Adapter~\cite{ye2023ip-adapter}. When used along with ControlNet, we employ depth maps and HED \cite{xie2015hed} images as the conditions, synthesizing visual contents prompted by the texts shown beneath the conditional images. As we can see, this combination produces quite acceptable outputs. When combined with IP-Adapter, we input a set of images featuring different objects. The results show that this combination not only generates the target IPs but also effectively characterizes the key components of the specified style.

\textbf{StyleBlend on more T2I diffusion models.}
Since the Stable Diffusion series \cite{Rombach2021HighResolutionIS, Podell2023SDXLIL} share a similar architecture, we can easily apply StyleBlend across different base models. For SDXL, which includes two text encoders, we train separate LoRAs on each to learn the composition style. Given SDXL's larger number of parameters, we increase the optimization steps to 400 for composition style learning and 600 for texture style learning. The configurations for SD v1.5 and v2.1 follow the same setup as described in Sec. \ref{sec:implement}. Fig. \ref{fig:variations} shows the creative results of StyleBlend on three Stable Diffusion models. By providing identical prompts and using the pretrained weights of StyleBlend for a specific style, our method can generate creative visual content that is both correct in semantics and consistent and coherent in style.

\subsection{Limitations and discussions}
\begin{figure}[t]
    \centering
    \includegraphics[width=1.0\linewidth]{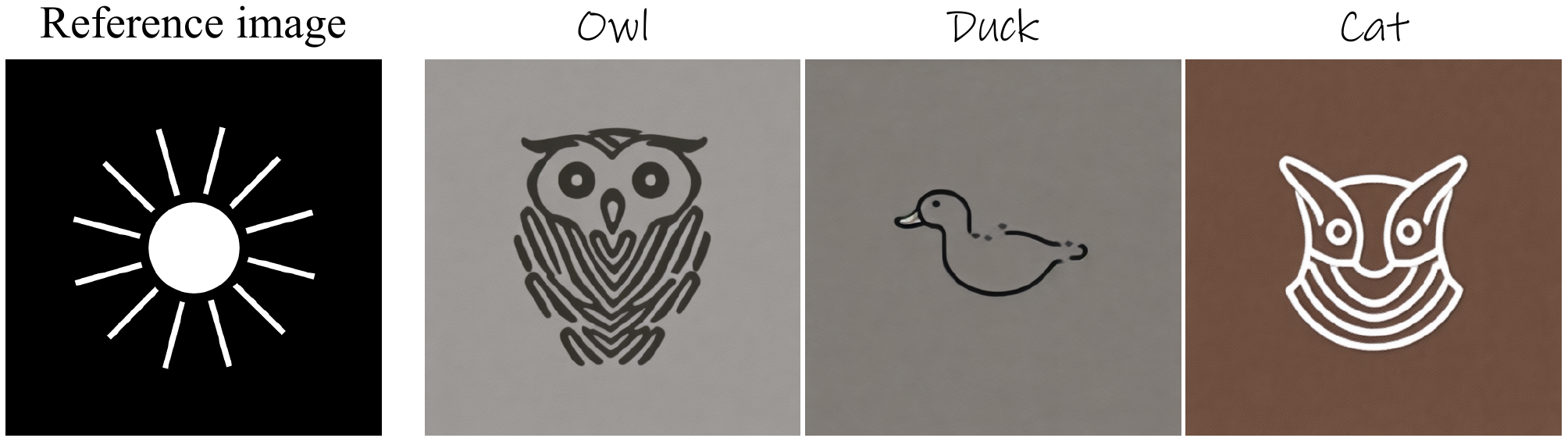}
    % \vspace*{-3mm}
        \caption{
    % \scriptsize
    Limitation of StyleBlend.}
    \label{fig:limitation}
    % \vspace*{-2mm}
\end{figure}

While we have demonstrated the effectiveness of StyleBlend across various style examples, several limitations exist. First, while our method performs well in 3-shot cases, it occasionally struggles with text-semantics alignment when using only one reference image, as reflected by the text alignment score in Fig. \ref{fig:comp_score}. Besides, for extreme cases shown in Fig. \ref{fig:limitation}, although our method captures the line style from the ``sun'' icon, it fails to replicate the black-and-white color accurately. Moreover, our current style-blending approach, which utilizes two branches for synthesis, is not the most efficient, as it doubles the inference cost. Future work may focus on developing a more efficient way to blend the two learned style components.

\section{Conclusions}
We have presented StyleBlend, an approach designed to enhance style-specific text-to-image generation based on the Stable Diffusion series. By employing two distinct training strategies to capture both the composition and texture styles of images using a limited set of reference images, StyleBlend effectively prepares for the blending of these styles in the synthesis process. The dual-branch synthesis framework integrates composition and texture styles to generate final outputs, achieving style coherence and text alignment in the visual contents. {Last but not least,} StyleBlend is efficient in terms of training time, the number of reference images required, and the number of learnable parameters.

\section*{Acknowledgments}
This work was supported in parts by NSFC (U21B2023),~Guangdong Basic and Applied Basic Research Foundation (2023B151
5120026), DEGP Innovation Team (2022KCXTD025), and Scientific Development Funds from Shenzhen University.

%-------------------------------------------------------------------------
% bibtex
%\bibliographystyle{eg-alpha-doi}  
%\bibliography{egbibsample}        

% biblatex with biber
\printbibliography                

%-------------------------------------------------------------------------
%Color tables are no longer required for purely electronic publications.
\newpage

% \begin{figure*}[tbp]
%   \centering
%   \mbox{} \hfill
%   % the following command controls the width of the embedded PS file
%   % (relative to the width of the current column)
%   \includegraphics[width=.3\linewidth]{sampleFig}
%   % replacing the above command with the one below will explicitly set
%   % the bounding box of the PS figure to the rectangle (xl,yl),(xh,yh).
%   % It will also prevent LaTeX from reading the PS file to determine
%   % the bounding box (i.e., it will speed up the compilation process)
%   % \includegraphics[width=.3\linewidth, bb=39 696 126 756]{sampleFig}
%   \hfill
%   \includegraphics[width=.3\linewidth]{sampleFig}
%   \hfill \mbox{}
%   \caption{\label{fig:ex3}%
%            For publications with color tables (i.e., publications not offering
%            color throughout the paper) please \textbf{observe}: 
%            for the printed version -- and ONLY for the printed
%            version -- color figures have to be placed in the last page.
%            \newline
%            For the electronic version, which will be converted to PDF before
%            making it available electronically, the color images should be
%            embedded within the document. Optionally, other multimedia
%            material may be attached to the electronic version. }
% \end{figure*}

\end{document}